\documentclass{article}
\pdfoutput=1
\usepackage{spconf,amsmath,amssymb,graphicx}
\usepackage{comment}
\usepackage{multirow}
\usepackage{enumitem}

\title{Extracting Causal Visual Features for Limited label Classification}
\name{Mohit Prabhushankar and Ghassan AlRegib}
\address{
School of Electrical and Computer Engineering,\\ Georgia Institute of Technology, Atlanta, GA, 30332-0250\\ \{mohit.p, alregib\}@gatech.edu}

\begin{document}

\onecolumn 

\begin{description}[leftmargin=2cm,style=multiline]

\item[\textbf{Citation}]{M. Prabhushankar, and G. AlRegib, "Extracting Causal Visual Features for Limited Label Classification," submitted to \emph{IEEE International Conference on Image Processing (ICIP)}, Sept. 2021.}

\item[\textbf{Copyright}]{\textcopyright 2020 IEEE. Personal use of this material is permitted. Permission from IEEE must be obtained for all other uses, in any current or future media, including reprinting/republishing this material for advertising or promotional purposes,
creating new collective works, for resale or redistribution to servers or lists, or reuse of any copyrighted component
of this work in other works. }

\item[\textbf{Contact}]{mohit.p@gatech.edu  OR alregib@gatech.edu\\ https://ghassanalregib.info/ \\ }
\end{description}

\thispagestyle{empty}
\newpage
\clearpage
\setcounter{page}{1}

\twocolumn

\ninept
\maketitle

\begin{abstract}
Neural networks trained to classify images do so by identifying features that allow them to distinguish between classes. These sets of features are either causal or context dependent. Grad-CAM is a popular method of visualizing both sets of features. In this paper, we formalize this feature divide and provide a methodology to extract causal features from Grad-CAM. We do so by defining context features as those features that allow \emph{contrast} between predicted class and any contrast class. We then apply a set theoretic approach to separate causal from contrast features for COVID-19 CT scans. We show that on average, the image regions with the proposed causal features require $15\%$ less bits when encoded using Huffman encoding, compared to Grad-CAM, for an average increase of $3\%$ classification accuracy, over Grad-CAM. Moreover, we validate the transfer-ability of causal features between networks and comment on the non-human interpretable causal nature of current networks.

\end{abstract}
\begin{keywords}
Visual Causality, Contrastive Explanations, COVID-19,  Gradients, Causal metrics
\end{keywords}

\section{Introduction}
\label{sec:intro}

In the field of image classification, deep learning networks have surpassed the top-5 human error rate~\cite{he2016deep} on ImageNet dataset~\cite{russakovsky2015imagenet}. In this dataset, neural networks learn to differentiate between $1000$ classes of natural images. The success of deep learning networks on natural images has fostered its usage on computed visual data including biomedical~\cite{temel2019relative} and seismic~\cite{alaudah2019machine, shafiq2018towards} fields. While the number of learn-able classes in these fields is generally limited, neural networks have the additional task of aiding domain specific experts to interpret the explanations behind their decisions to promote trust in the network. For instance, in the field of biomedical imaging, a medical practitioner diagnoses whether a patient is COVID positive or negative based on CT scans~\cite{zhao2020covid}. The authors in~\cite{he2020sample} use transfer learning approaches on CT scans to perform the detection and provide explanatory results using Grad-CAM~\cite{selvaraju2017grad} to justify their network's efficacy in detecting COVID-19. Grad-CAM highlights features in the image that lead to the network's decision. In this paper, we analyze a neural network's causal capability using existing explanatory methods by providing a technique to extract causal features from such explanatory methods.

Probabilistic causation assumes a causal relationship between two events $C$ and $P$ if event $C$ increases the probability of occurrence of $P$~\cite{hitchcock1997probabilistic}. In image classification networks, $P$ refers to decisions made by neural networks based on features $C$. A popular method for ascertaining probabilistic causality is through interventions in data~\cite{pearl2000models}. In these models, the set of causal features $C$ is varied by intervening in the generation process of $C$ to ascertain the change in the observed decision $P$. Such interventions can however be long, complex, unethical or impossible~\cite{steyvers2003inferring} like in COVID CT scans. Hence, we forego interventionist causality and rely on \emph{observed causality} to derive causation. Observed causality relies on passive observation to determine statistical causality. The authors in~\cite{lopez2017discovering} propose that non-interventionist observation provides two sets of features - causal $C$, and context $B$ - that lead to decision $P$. In other words, a decision $P$ is made based on both causal and context features in an image. Hence, existing explanatory methods including~\cite{selvaraju2017grad,chattopadhay2018grad,springenberg2014striving} highlight $C \cup B$ features. However, they do not provide a methodology to extract either $C$ or $B$ separately. In this paper, we utilize contrastive features from~\cite{prabhushankar2020contrastive} to approximate $B$ features. We then propose a set-theoretic approach to abstract $C$ out of Grad-CAM's $C \cup B$ features. The contributions of this paper include:

\begin{itemize}
    \item Formulating a set theoretic interpretation of causal and context features for observed causality in visual data.
    \item Expressing context features as contrastive features.
    \item Providing an evaluation setup that tests for causality in a limited label scenario.
\end{itemize}

In Section~\ref{sec:Lit_Review}, we motivate context via contrast and review Grad-CAM and contrastive explanations. We then motivate the proposed method and detail its procedure in Section~\ref{sec:Method}. We finally present the results in Section~\ref{sec:Experiments} before concluding in Section~\ref{sec:Conclusion}.
\begin{figure*}[!htb]
\begin{center}
\minipage{1\textwidth}%
  \includegraphics[width=\linewidth]{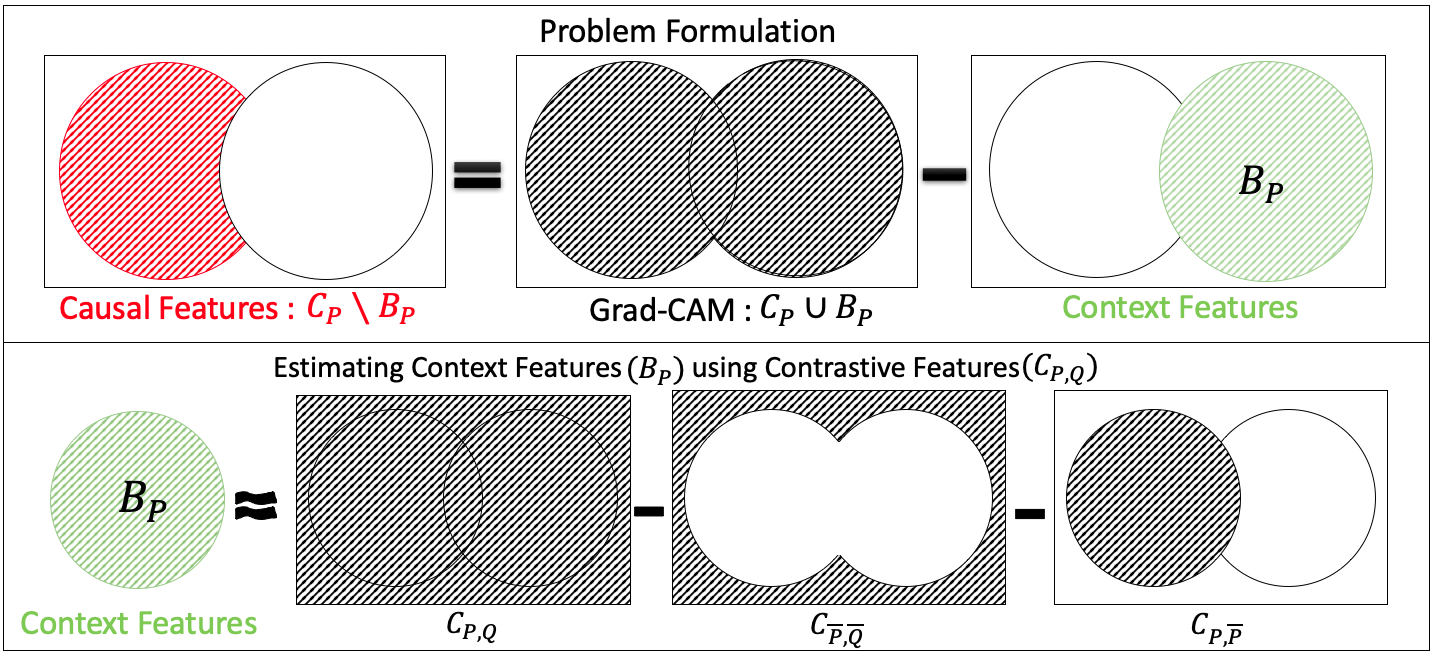}
\endminipage
\caption{Top : Venn diagram for problem formulation based on Eq.~\ref{eq:required}. Bottom : Estimating context features from Eq.~\ref{eq:final_set}. Note that because the network does not classify with $100\%$ confidence, we cannot resolve $C_P \cap B_P$.}\label{fig:Venn}\vspace{-5mm}
\end{center}
\end{figure*}

\section{Background and Related Works}
\label{sec:Lit_Review}

\noindent\textbf{Causal and Context features: }The authors in~\cite{lopez2017discovering} define causal features as visual features that exist within the physical body of the object in an image and context features as visual features that surround the object in the image. In this paper, we forego the definitions based on physical locations in favor of the feature's membership towards predicting a class $P$. We define causal features $C$ as those features whose presence increases the likelihood of occurrence of decision $P$ in any CT scan $x$. Conversely, the absence of causal features $C$ decreases the probability of decision $P$. The above two definitions of causal features are derived from the Common Cause Principles~\cite{hofer1999reichenbach} and are used in~\cite{petsiuk2018rise} to evaluate causality. We follow a slightly altered methodology to showcase the causal effectiveness of our method. We define context features $B$ contrastively, as features that allow differentiating predicted class $P$ and a contrast class $Q$, without necessarily causing $P$.

\noindent\textbf{Context and Contrast features: }In the field of human visual saliency, the authors in~\cite{oliva2007role} provide an argument for the existence of contextual features of a class that are represented by their relationship with features of other classes. In~\cite{sun2020implicit}, the implicit saliency of a neural network is extracted as an expectancy-mismatch between the predicted class against all learned classes thereby empirically validating the existence of \emph{contrastive} information within neural networks. The authors in~\cite{prabhushankar2020contrastive} extract this information and visualize them as explanations. In this paper, we represent the context features $B$ as contrast features. 

\noindent\textbf{Grad-CAM and Contrastive Explanations: }Consider a trained binary classifier $f()$. Given an input image $x$, $y = f(x)$ are the logit outputs of dimensions $2\times 1$. The predicted class $P$ of image $x$ is the index of the maximum element in $y$ i.e. $P = \operatorname{arg\,max}_i  y_i,  \forall i \in [1,2]$. Grad-CAM localizes all features in $x$ that leads to a decision $P$ by backpropagating the logit $y_P$ to the last convolutional layer $l$. The per-channel gradients in layer $l$ are summed up to obtain an importance score $\alpha_k$ for a channel $k, k \in [1,K]$ and multiplied with the activations in their respective channels $A_k$. The  importance score weighted activation maps are averaged to obtain the Grad-CAM mask $\mathcal{G}_P = ReLU(\sum_{k=1}^K \alpha_k A^k)$ for class $P$. The authors in~\cite{prabhushankar2020contrastive} modified the Grad-CAM framework to backpropagate a loss function $\mathcal{L}_{P,Q}$ between predicted class $P$ and a contrast class $Q$. With the other steps remaining the same, a contrast-importance score $\alpha_k^c$ weighted contrast mask is given by $\mathcal{C} = ReLU(\sum_{k=1}^K \alpha_k^c A^k)$ for predicted and contrast classes $P$ and $Q$. Note that gradients are used as features in multiple works including~\cite{kwon2019distorted, kwon2020backpropagated, lee2020gradients}.

\begin{figure*}[!htb]
\begin{center}
\minipage{1\textwidth}%
  \includegraphics[width=\linewidth]{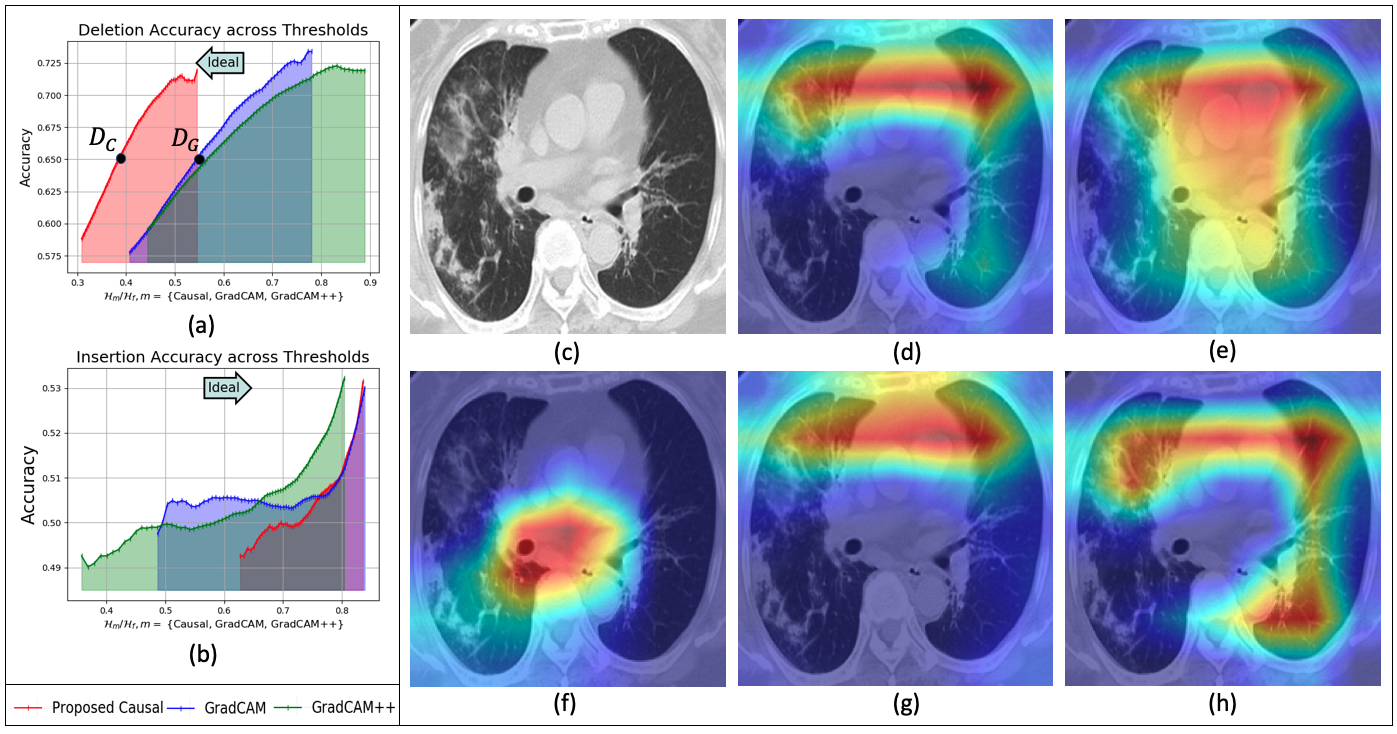}
\endminipage
\caption{(a) Deletion - Curves to the left are ideal. (b) Insertion - Curves to the right are ideal. (c) Original scan. (d) Grad-CAM. (e) Grad-CAM++. (f) Proposed causal explanation. (g) $\mathcal{C_{P,Q}}$. (h) $\mathcal{C_{\overline{P},\overline{P}}}$}\label{fig:Results}\vspace{-5mm}
\end{center}
\end{figure*}
\vspace{-3mm}
\section{Proposed Method}
\label{sec:Method}
\vspace{-2mm}
We first motivate our method based on set theory before describing the process of extraction of causal features.
\vspace{-2mm}
\subsection{Theory}
Consider the setting as described in Section~\ref{sec:Lit_Review} where a binary classification network $f()$ is trained on COVID-19 CT scans~\cite{zhao2020covid}. Once trained, for any given scan from the dataset, Grad-CAM provides visual features that combine both causal and context features. Hence, Grad-CAM provides a mask, $\mathcal{G}_P = C_P \cup B_P$ for the prediction $P$ on a given scan $x$. If the network classifies $x$ correctly with $100\%$ confidence, then $f()$ has resolved causal and context features independently such that $\mathcal{G}_P = C_P + B_P$. However, this rarely occurs in practice and we assume $\mathcal{G}_P = C_P + B_P - (C_P \cap B_P)$. Hence, our goal is to extract the relative complement $C_P \setminus B_P$ given $\mathcal{G}_P = C_P \cup B_P$. This is illustrated in Fig.~\ref{fig:Venn}. Based on a visual inspection of the venn diagram, we can rewrite $C_P \setminus B_P$ as,
\begin{equation}\label{eq:required}
    C_P \setminus B_P = \mathcal{G}_P - B_P.
\end{equation}
Note that we do not have access to either $C_P$ or $B_P$. We are only provided with $\mathcal{G}_P$. In this paper, we estimate the context features $B_P$ using contrastive features from~\cite{prabhushankar2020contrastive}. Specifically, continuing the notations from Section~\ref{sec:Lit_Review}, we represent $B_P$ as,
\begin{align}\label{eq:final_set}
    B_P &= \mathcal{C_{P,Q}} - \mathcal{C}_{\overline{P},\overline{Q}} - \mathcal{C_{P,\overline{P}}}.
\end{align}
Substituting Eq.~\ref{eq:final_set} back in Eq.~\ref{eq:required}, we obtain our final formulation,
\begin{align}\label{eq:final}
    C_P \setminus B_P &= \mathcal{G}_P - [\mathcal{C_{P,Q}} - \mathcal{C}_{\overline{P},\overline{Q}} - \mathcal{C_{P,\overline{P}}}].
\end{align}
A venn diagram visualization is presented in Fig.~\ref{fig:Venn}. We qualitatively explain all the contrastive terms.

\subsection{Contrastive features}

\noindent\textbf{$\mathcal{C_{P,Q}}$: }Highlights features that answer \emph{`Why P or Q?'}. This term contrastively leads to either decisions of $P$ or $Q$. In the binary setting, we approximate this to be all possible features $\mathcal{U}$. Borrowing notations from Section~\ref{sec:Lit_Review}, $\mathcal{C_{P,Q}}$ is obtained by backpropagating a loss $\mathcal{L}(y, [1,1])$ to obtain a contrast-importance score $\alpha_k^{P,Q}$.

\textbf{$\mathcal{C}_{\overline{P},\overline{Q}}$: }Highlights features that answer \emph{`Why neither P nor Q?'}. The features in this term do not increase the probability of either $P$ or $Q$. $\mathcal{C_{\overline{P},\overline{Q}}}$ is obtained by backpropagating a loss $\mathcal{L}(y, [0,0])$ to obtain a contrast-importance score $\alpha_k^{\overline{P},\overline{Q}}$.

\textbf{$\mathcal{C_{P,\overline{P}}}$: }Highlights features that answer \emph{`Why not P with 100\% confidence?'}. Hence, it highlights all unresolved causal features. $\mathcal{C_{P,\overline{P}}}$ is obtained by backpropagating a loss $\mathcal{L}(y, [1,0])$ to obtain a contrast-importance score $\alpha_k^{P,\overline{P}}$

\subsection{Implementation}
Continuing the notations from Section~\ref{sec:Lit_Review}, the implementation equivalent of Eq.~\ref{eq:final} is given by,
\begin{equation}\label{eq:gradient}
    C_P \setminus B_P =  ReLU\bigg(\sum_{k=1}^K -\bigg[\alpha_k - \alpha_k^{P,Q} + \alpha_k^{\overline{P}, \overline{Q}} + \alpha_k^{P, \overline{P}}\bigg] A^k\bigg),
\end{equation}
where $\alpha_k$ represents the importance score from Grad-CAM and $\alpha_k^c$ represents the normalized importance score from contrast maps. The overall negative sign occurs because $\alpha$ are gradients whose directions are opposite to the feature minima. The final map is normalized and is visualized. A representative COVID negative scan and its Grad-CAM~\cite{selvaraju2017grad} and Grad-CAM++~\cite{chattopadhay2018grad} explanations are shown in Figs.~\ref{fig:Results}c, ~\ref{fig:Results}d, and~\ref{fig:Results}e respectively. The causal map from Eq.~\ref{eq:gradient} and contrastive maps $\mathcal{C_{P,Q}}$ and $\mathcal{C_{\overline{P},\overline{P}}}$ are visualized in Figs.~\ref{fig:Results}f, ~\ref{fig:Results}g and~\ref{fig:Results}h respectively. Note that while $\mathcal{C_{P,Q}}$ appears similar to $\mathcal{G}$, $\mathcal{C_{P,Q}}$ is biased by normalization and its $\alpha_k$ values are lesser.

\noindent\textbf{Effect of number of classes: }In a binary classification setting, we need four feature maps - one Grad-CAM and three contrastive maps to extract causal features. These are obtained by backpropagating $\{(0,0), (1,0), (1,1)$ and the logit for Grad-CAM. Hence, we backpropagate the power set of all possible class combinations. This translates to $2^N$ backpropagations for $N$ classes. Therefore, this technique is suitable for a limited class scenario.

\vspace{-3mm}
\section{Experiments}
\label{sec:Experiments}
\vspace{-3mm}
In this section, we detail the experiments to validate the causal nature of our proposed features. We perform two sets of experiments to validate within-network and inter-network causality. The COVID-19 dataset~\cite{zhao2020covid} consists of 349 COVID positive CT scans and 463 COVID negative CT scans. We train ResNets-18,34,50~\cite{he2016deep} and DenseNets-121,169~\cite{huang2017densely} as described in~\cite{he2020sample}.
\vspace{-2mm}
\subsection{Within-network causality : Deletion and Insertion}\label{subsec:Del}
\vspace{-2mm}
The authors in~\cite{petsiuk2018rise} propose two causal metrics - deletion and insertion. In deletion, the identified causes are deleted pixel by pixel and the probability of predicted class, as a function of the fraction of the removed pixels, is monitored. In insertion, the non-causal pixels are added and the increase in probability as a function of added fraction of pixels is noted. However, in a binary setting, the probability for a class rarely decreases to a large extent even after removing a majority of the pixels. Hence, we modify the deletion and insertion setup to measure accuracy instead of probability on masked images.

\begin{figure}[!htb]
\begin{center}
\minipage{0.48\textwidth}%
  \includegraphics[width=\linewidth]{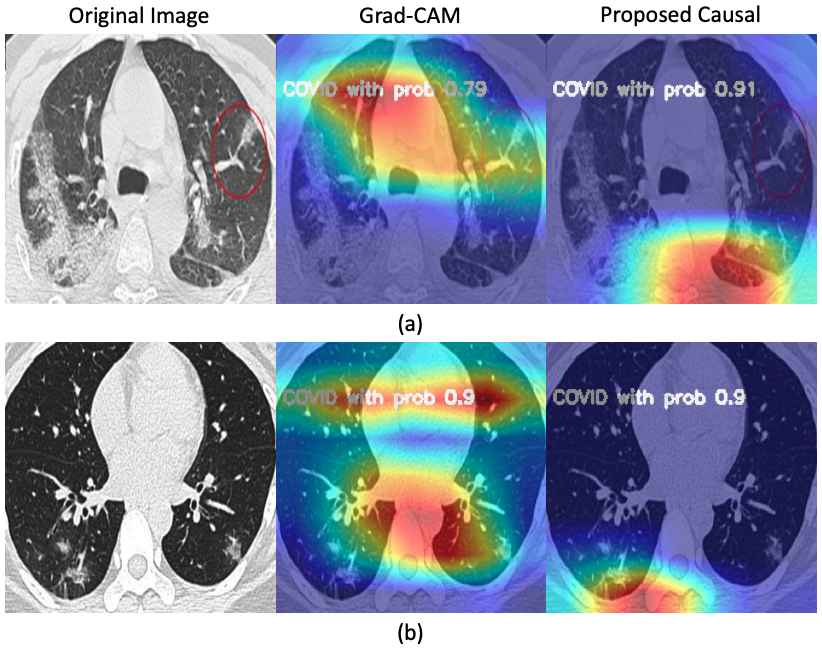}
\endminipage
\caption{(a) Non human-interpretable causal feature has higher prediction confidence. (b) The prediction confidences from both explanations are equal.}\label{fig:Qualitative}\vspace{-9mm}
\end{center}
\end{figure}

A threshold is applied on the Grad-CAM~\cite{selvaraju2017grad}, Grad-CAM++~\cite{chattopadhay2018grad} and proposed causal maps. For deletion, the pixels greater than the given threshold are made equal to $1$ with the rest being $0$. And vice-versa for insertion. The binary mask is then multiplied with the original input image and the masked image is passed through the model. The model's prediction is noted. This is conducted on all images in the testing set and the average test accuracy is calculated. The experiment is conducted with $81$ thresholds ranging from $0.1$ to $0.8$ with an increment of $0.01$. The average accuracy in each case is noted. From Figs.~\ref{fig:Results}f and~\ref{fig:Qualitative}, we see that the area highlighted by the proposed causal features is lesser than the compared methods. To objectively measure this area, we encode the original and masked images using Huffman coding~\cite{huffman1952method} as $\mathcal{H}_f$ and $\mathcal{H}_m$ respectively. The ratio of the bits is taken as $\mathcal{H} = \mathcal{H}_m/\mathcal{H}_f$. Each average accuracy for a threshold is now associated with an $\mathcal{H}$. All $81$ accuracies are plotted as a function of their $\mathcal{H}$ and depicted in Fig.~\ref{fig:Results}a. Consider two points $D_C$ and $D_{\mathcal{G}}$ on the proposed causal and Grad-CAM curves respectively. These points depict roughly $65\%$ averaged accuracy. From the corresponding bit rates in the $\texttt{x-axis}$, the causal features achieve this accuracy at a lower bit rate compared to Grad-CAM. Hence, dense causal features are encoded by lesser bits in the proposed method. This is validated in the insertion plot in Fig.~\ref{fig:Results}b as well.
\begin{table*}[tb]
\small
\centering
\caption{Causal Feature Transference from ResNet-18 to other architectures.\label{tab:Res-18}}
\begin{tabular}{ccccccccccc}
\hline\hline
\multirow{3}{*}{Threshold} & \multicolumn{2}{c}{\multirow{2}{*}{Huffman ($\downarrow$)}} & \multicolumn{8}{c}{Accuracies ($\uparrow$)} \\ \cline{4-11} 
& & & \multicolumn{2}{c}{ResNet-34} & \multicolumn{2}{c}{ResNet-50} & \multicolumn{2}{c}{DenseNet-121} & \multicolumn{2}{c}{DenseNet-169} \\ \cline{2-11}
&  GradCAM & Causal & GradCAM & Causal & GradCAM & Causal & GradCAM & Causal & GradCAM & Causal \\

\hline\hline

0.1 & 0.7802 & \textbf{0.5456} & 0.6158 & \textbf{0.6502} & \textbf{0.7586} & 0.7537 & 0.6404 & \textbf{0.6453} & 0.7044 & \textbf{0.7291} \\

0.2 & 0.6442 & \textbf{0.4549} & 0.5911 & \textbf{0.6355} & 0.7734 & \textbf{0.7783} & 0.6158 & \textbf{0.6256} & 0.7143 & \textbf{0.7685} \\

0.3 & 0.5329 & \textbf{0.3879} & 0.5665 & \textbf{0.5764} & 0.7241 & \textbf{0.7980} & 0.6108 & \textbf{0.6207} & 0.6946 & \textbf{0.7389} \\

0.4 & 0.4434 & \textbf{0.3329} & 0.5074 & \textbf{0.5419} & 0.67 & \textbf{0.7882} & 0.5911 & \textbf{0.5961} & 0.6305 & \textbf{0.7192} \\

0.5 & 0.3715 & \textbf{0.2886} & 0.5025 & \textbf{0.5222} & 0.601 & \textbf{0.7586} & 0.5911 & \textbf{0.6108} & 0.6059 & \textbf{0.6847} \\

\hline\hline
\end{tabular}
\end{table*}

\begin{table*}[tb]
\small
\centering
\vspace{-5mm}\caption{Causal Feature Transference from ResNet-34 to other architectures.\label{tab:Res-34}}
\begin{tabular}{ccccccccccc}
\hline\hline
\multirow{3}{*}{Threshold} & \multicolumn{2}{c}{\multirow{2}{*}{Huffman ($\downarrow$)}} & \multicolumn{8}{c}{Accuracies ($\uparrow$)} \\ \cline{4-11} 
& & & \multicolumn{2}{c}{ResNet-18} & \multicolumn{2}{c}{ResNet-50} & \multicolumn{2}{c}{DenseNet-121} & \multicolumn{2}{c}{DenseNet-169} \\ \cline{2-11}
&  GradCAM & Causal & GradCAM & Causal & GradCAM & Causal & GradCAM & Causal & GradCAM & Causal \\

\hline\hline

0.1 & 0.8352  & \textbf{0.6531} & \textbf{0.7094} & 0.7044 & \textbf{0.7783} & 0.7241  & 0.6108 & \textbf{0.6552} & 0.7389 & \textbf{0.7586} \\

0.2 & 0.7493 & \textbf{0.5646} & 0.7044 & \textbf{0.6995} & \textbf{0.7931} & 0.7586 & 0.6059 & \textbf{0.6256} & 0.7537 & \textbf{0.7586} \\

0.3 & 0.6584 & \textbf{0.4781} & \textbf{0.6749} & \textbf{0.6749} & \textbf{0.8177} & 0.7537 & \textbf{0.6059} & \textbf{0.6059} & \textbf{0.7389} & 0.7340 \\

0.4 & 0.5672 & \textbf{0.3983} & 0.6502 & \textbf{0.6650} & 0.7635 & \textbf{0.7685} & \textbf{0.6059} & 0.5911 & \textbf{0.7192} & 0.7044 \\

0.5 & 0.4749 & \textbf{0.3292} & 0.6010 & \textbf{0.6059} & \textbf{0.7783} & 0.7537 & \textbf{0.5764} & 0.5616 & \textbf{0.6897} & 0.6552 \\

\hline\hline
\end{tabular}
\end{table*}
\vspace{-2mm}
\subsection{Inter-network causality : Transference of features}\label{subsec:Trans}
\vspace{-2mm}
In this section, we mask input images based on features obtained from the proposed causal and Grad-CAM methods using ResNet-18~\cite{he2016deep}. We then pass these masked images through other trained networks including ResNets-34,50~\cite{he2016deep} and DenseNets-121,169~\cite{huang2017densely}. This experiment is designed to validate the transfer-ability of causal features identified by ResNet-18 to other networks. The accuracy and Huffman ratio results for $5$ different thresholds are shown in Table~\ref{tab:Res-18}. It can be seen that the huffman ratio for the proposed method is lesser than Grad-CAM for all thresholds. Hence, it is able to identify dense causal features from Grad-CAM. The averaged accuracy of masked images is also shown for $4$ other networks. In $19$ of the $20$ categories, the proposed causal feature masked images outperform Grad-CAM feature masked images with a lesser huffman ratio. In Table~\ref{tab:Res-34}, we extract masks using ResNet-34, perform deletion based on shown thresholds and obtain huffman ratios for all test images. These masked images are then passed into the corresponding networks and the accuracy results are shown. In $10$ of the $20$ categories, the proposed causal features outperform Grad-CAM features.

\vspace{-2mm}
\subsection{Qualitative Analysis}
\vspace{-2mm}
The authors in~\cite{petsiuk2018rise} argue that humans must be kept out of the loop when evaluating causality. However, by definition, explanations are rationales used by networks to justify their decisions~\cite{kitcher1962scientific}. These justifications are made for the benefit of humans. Such justifications are required in fields like biomedical imaging where deep learning tools are used as aids by medical practitioners. We visualize Grad-CAM and their underlying causal features from the proposed technique in Fig.~\ref{fig:Qualitative}. Both original scans are from COVID positive patients. In Fig.~\ref{fig:Qualitative}a, Grad-CAM fails to highlight the circled red region that depicts COVID. More importantly, the extracted causal features are at the bottom right. Feeding the masked image into ResNet-18, the network classifies both correctly but with a higher confidence in the causal features. In Fig.~\ref{fig:Qualitative}b, we pick a scan whose Grad-CAM and causal features were classified with the same confidence but from different regions within the scans. 

These results suggest that it is the context features that add human interpretability and causal features that aid classification. In real-world biomedical applications like in the considered COVID-19 detection, it is imperative to identify and make decisions based on causal features. It merits further study into designing better networks whose causal features are more human interpretable, similar to Grad-CAM's causal and context feature set.

\vspace{-2mm}
\section{Conclusion}
\label{sec:Conclusion}
\vspace{-2mm}
In this paper we formalize the causal and context features that a neural network bases its decision on. We express context features in terms of contrastive features between classes that the neural network has implicitly learned. This allows separation between causal and context features. Grad-CAM is used as the explanatory mechanism from which causal features are extracted. We validate and establish the trasfer-abilty of these causal features across networks. The visualizations suggest that the causal regions that a neural network bases its decision on is not always human interpretable. This calls for more work in designing human-interpretable causal features especially in fields like biomedical imaging.

\bibliographystyle{IEEEbib}
\bibliography{refs}

\begin{thebibliography}{10}

\bibitem{he2016deep}
Kaiming He, Xiangyu Zhang, Shaoqing Ren, and Jian Sun,
\newblock ``Deep residual learning for image recognition,''
\newblock in {\em Proceedings of the IEEE conference on computer vision and
  pattern recognition}, 2016, pp. 770--778.

\bibitem{russakovsky2015imagenet}
Olga Russakovsky, Jia Deng, Hao Su, Jonathan Krause, Sanjeev Satheesh, Sean Ma,
  Zhiheng Huang, Andrej Karpathy, Aditya Khosla, Michael Bernstein, et~al.,
\newblock ``Imagenet large scale visual recognition challenge,''
\newblock {\em International journal of computer vision}, vol. 115, no. 3, pp.
  211--252, 2015.

\bibitem{temel2019relative}
Dogancan Temel, Melvin~J Mathew, Ghassan AlRegib, and Yousuf~M Khalifa,
\newblock ``Relative afferent pupillary defect screening through transfer
  learning,''
\newblock {\em IEEE Journal of Biomedical and Health Informatics}, vol. 24, no.
  3, pp. 788--795, 2019.

\bibitem{alaudah2019machine}
Yazeed Alaudah, Patrycja Micha{\l}owicz, Motaz Alfarraj, and Ghassan AlRegib,
\newblock ``A machine-learning benchmark for facies classification,''
\newblock {\em Interpretation}, vol. 7, no. 3, pp. SE175--SE187, 2019.

\bibitem{shafiq2018towards}
Muhammad~A Shafiq, Mohit Prabhushankar, Haibin Di, and Ghassan AlRegib,
\newblock ``Towards understanding common features between natural and seismic
  images,''
\newblock in {\em SEG Technical Program Expanded Abstracts 2018}, pp.
  2076--2080. Society of Exploration Geophysicists, 2018.

\bibitem{zhao2020covid}
Jinyu Zhao, Yichen Zhang, Xuehai He, and Pengtao Xie,
\newblock ``Covid-ct-dataset: a ct scan dataset about covid-19,''
\newblock {\em arXiv preprint arXiv:2003.13865}, 2020.

\bibitem{he2020sample}
Xuehai He, Xingyi Yang, Shanghang Zhang, Jinyu Zhao, Yichen Zhang, Eric Xing,
  and Pengtao Xie,
\newblock ``Sample-efficient deep learning for covid-19 diagnosis based on ct
  scans,''
\newblock {\em medRxiv}, 2020.

\bibitem{selvaraju2017grad}
Ramprasaath~R Selvaraju, Michael Cogswell, Abhishek Das, Ramakrishna Vedantam,
  Devi Parikh, and Dhruv Batra,
\newblock ``Grad-cam: Visual explanations from deep networks via gradient-based
  localization,''
\newblock in {\em Proceedings of the IEEE International Conference on Computer
  Vision}, 2017, pp. 618--626.

\bibitem{hitchcock1997probabilistic}
Christopher Hitchcock,
\newblock ``Probabilistic causation,''
\newblock 1997.

\bibitem{pearl2000models}
Judea Pearl et~al.,
\newblock ``Models, reasoning and inference,''
\newblock {\em Cambridge, UK: CambridgeUniversityPress}, 2000.

\bibitem{steyvers2003inferring}
Mark Steyvers, Joshua~B Tenenbaum, Eric-Jan Wagenmakers, and Ben Blum,
\newblock ``Inferring causal networks from observations and interventions,''
\newblock {\em Cognitive science}, vol. 27, no. 3, pp. 453--489, 2003.

\bibitem{lopez2017discovering}
David Lopez-Paz, Robert Nishihara, Soumith Chintala, Bernhard Scholkopf, and
  L{\'e}on Bottou,
\newblock ``Discovering causal signals in images,''
\newblock in {\em Proceedings of the IEEE Conference on Computer Vision and
  Pattern Recognition}, 2017, pp. 6979--6987.

\bibitem{chattopadhay2018grad}
Aditya Chattopadhay, Anirban Sarkar, Prantik Howlader, and Vineeth~N
  Balasubramanian,
\newblock ``Grad-cam++: Generalized gradient-based visual explanations for deep
  convolutional networks,''
\newblock in {\em 2018 IEEE Winter Conference on Applications of Computer
  Vision (WACV)}. IEEE, 2018, pp. 839--847.

\bibitem{springenberg2014striving}
Jost~Tobias Springenberg, Alexey Dosovitskiy, Thomas Brox, and Martin
  Riedmiller,
\newblock ``Striving for simplicity: The all convolutional net,''
\newblock {\em arXiv preprint arXiv:1412.6806}, 2014.

\bibitem{prabhushankar2020contrastive}
Mohit Prabhushankar, Gukyeong Kwon, Dogancan Temel, and Ghassan AlRegib,
\newblock ``Contrastive explanations in neural networks,''
\newblock in {\em 2020 IEEE International Conference on Image Processing
  (ICIP)}. IEEE, 2020, pp. 3289--3293.

\bibitem{hofer1999reichenbach}
G{\'a}bor Hofer-Szab{\'o}, Mikl{\'o}s R{\'e}dei, and L{\'a}szl{\'o}~E
  Szab{\'o},
\newblock ``On reichenbach's common cause principle and reichenbach's notion of
  common cause,''
\newblock {\em The British Journal for the Philosophy of Science}, vol. 50, no.
  3, pp. 377--399, 1999.

\bibitem{petsiuk2018rise}
Vitali Petsiuk, Abir Das, and Kate Saenko,
\newblock ``Rise: Randomized input sampling for explanation of black-box
  models,''
\newblock {\em arXiv preprint arXiv:1806.07421}, 2018.

\bibitem{oliva2007role}
Aude Oliva and Antonio Torralba,
\newblock ``The role of context in object recognition,''
\newblock {\em Trends in cognitive sciences}, vol. 11, no. 12, pp. 520--527,
  2007.

\bibitem{sun2020implicit}
Yutong Sun, Mohit Prabhushankar, and Ghassan AlRegib,
\newblock ``Implicit saliency in deep neural networks,''
\newblock in {\em 2020 IEEE International Conference on Image Processing
  (ICIP)}. IEEE, 2020, pp. 2915--2919.

\bibitem{kwon2019distorted}
Gukyeong Kwon, Mohit Prabhushankar, Dogancan Temel, and Ghassan AlRegib,
\newblock ``Distorted representation space characterization through
  backpropagated gradients,''
\newblock in {\em 2019 IEEE International Conference on Image Processing
  (ICIP)}. IEEE, 2019, pp. 2651--2655.

\bibitem{kwon2020backpropagated}
Gukyeong Kwon, Mohit Prabhushankar, Dogancan Temel, and Ghassan AlRegib,
\newblock ``Backpropagated gradient representations for anomaly detection,''
\newblock in {\em European Conference on Computer Vision}. Springer, 2020, pp.
  206--226.

\bibitem{lee2020gradients}
Jinsol Lee and Ghassan AlRegib,
\newblock ``Gradients as a measure of uncertainty in neural networks,''
\newblock in {\em 2020 IEEE International Conference on Image Processing
  (ICIP)}. IEEE, 2020, pp. 2416--2420.

\bibitem{huang2017densely}
Gao Huang, Zhuang Liu, Laurens Van Der~Maaten, and Kilian~Q Weinberger,
\newblock ``Densely connected convolutional networks,''
\newblock in {\em Proceedings of the IEEE conference on computer vision and
  pattern recognition}, 2017, pp. 4700--4708.

\bibitem{huffman1952method}
David~A Huffman,
\newblock ``A method for the construction of minimum-redundancy codes,''
\newblock {\em Proceedings of the IRE}, vol. 40, no. 9, pp. 1098--1101, 1952.

\bibitem{kitcher1962scientific}
Philip Kitcher and Wesley~C Salmon,
\newblock {\em Scientific explanation}, vol.~13,
\newblock U of Minnesota Press, 1962.

\end{thebibliography}

\end{document}